\documentclass[10pt,twocolumn,letterpaper]{article}

\usepackage{cvpr}              % To produce the CAMERA-READY version
\usepackage[utf8]{inputenc} % allow utf-8 input
\usepackage[T1]{fontenc}    % use 8-bit T1 fonts
\usepackage{url}            % simple URL typesetting
\usepackage{booktabs}       % professional-quality tables
\usepackage{amsfonts}       % blackboard math symbols
\usepackage{nicefrac}       % compact symbols for 1/2, etc.
\usepackage{microtype}      % microtypography
\usepackage{graphicx}
\usepackage{wrapfig}
\usepackage{algorithm,algpseudocode}
\usepackage{listings}
\usepackage[toc,page]{appendix}
\usepackage{colortbl} % rowcolor

\newcommand{\methodname}{3DHM}
\newcommand{\sonemethod}{Inpainting Diffusion}

\newcommand{\stwomethod}{Rendering Diffusion}

\definecolor{bleudefrance}{rgb}{0.0, 0.36, 1.0} 
\definecolor{lightblue}{rgb}{0.22,0.45,0.70}% light blue

\definecolor{backcolour}{RGB}{250,250,250}
\definecolor{keywords}{RGB}{255,0,90} 
\definecolor{comments}{RGB}{0,0,113}
\definecolor{codered}{RGB}{160,0,0}
\definecolor{codegreen}{RGB}{0,150,0}
\definecolor{codeblue}{RGB}{0,90,255}
\definecolor{codeorange}{RGB}{255,90,0}
\definecolor{codeorange}{RGB}{255,90,0}
\definecolor{aliceblue}{rgb}{0.91, 0.94, 0.97}

\lstdefinestyle{mypython}{language=Python, 
    basicstyle=\ttfamily\scriptsize, 
    backgroundcolor=\color{backcolour}, 
    keywordstyle=\color{codeorange},
    commentstyle=\color{comments},
    stringstyle=\color{codered},
    showstringspaces=false,
    identifierstyle=\color{black},      
    captionpos=b,
    escapechar={|}
}

\definecolor{cvprblue}{rgb}{0.21,0.49,0.74}
\usepackage[pagebackref,breaklinks,color links,citecolor=cvprblue]{hyperref}

\def\paperID{8684} % *** Enter the Paper ID here
\def\confName{CVPR}
\def\confYear{2025}

\title{Synthesizing Moving People with 3D Control}

\author{Boyi Li*
\quad
Junming Leo Chen*
\quad
Jathushan Rajasegaran*
\quad
Yossi Gandelsman \\
Alexei A. Efros
\quad
Jitendra Malik\\
UC Berkeley
}

\begin{document}

\twocolumn[{
\renewcommand\twocolumn[1][]{#1}

\maketitle

\begin{center}
    \centering
    % \vspace{-2em}
    \includegraphics[width=0.95\linewidth]{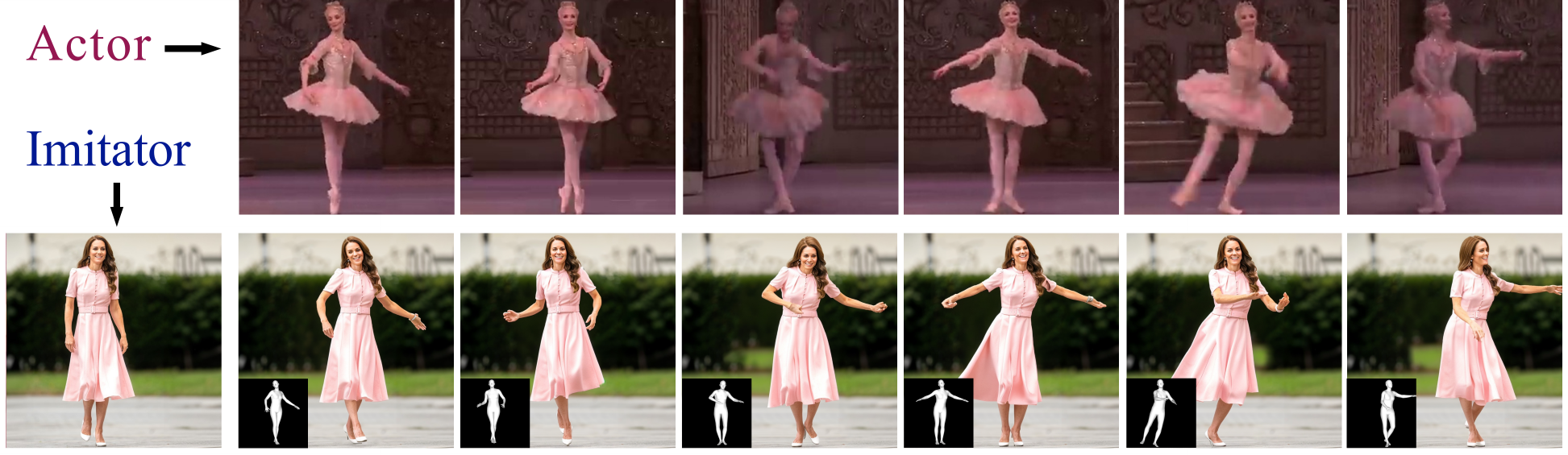}
    % \vspace{-2em}
    \captionof{figure}{\textbf{The Imitation Game:} Given a video of a person \textbf{"The Actor"}, we want to transfer their motion to a new person \textbf{"The Imitator"}. In this figure, the first row shows a sequence of frames of the actor from a ballerina \textit{Dance of the Sugar Plum Fairy}. The inset row shows the 3D poses extracted from this video. Now, given any single image of a new person  \textbf{The Imitator}, our model can synthesize new renderings of the imitator, to copy the actions of the actor in 3D.
    }
    \label{fig:teaser}
\end{center}

\vspace{-0.05in}
}]

\begin{abstract}
    In this paper, we present a diffusion model-based framework for animating people from a single image for a given target 3D motion sequence. Our approach has two core components: a) learning priors about invisible parts of the human body and clothing, and b) rendering novel body poses with proper clothing and texture. For the first part, we learn an in-filling diffusion model to hallucinate unseen parts of a person given a single image. We train this model on texture map space, which makes it more sample-efficient since it is invariant to pose and viewpoint. Second, we develop a diffusion-based rendering pipeline, which is controlled by 3D human poses. This produces realistic renderings of novel poses of the person, including clothing, hair, and plausible in-filling of unseen regions. This disentangled approach allows our method to generate a sequence of images that are faithful to the target motion in the 3D pose and, to the input image in terms of visual similarity. In addition to that, the 3D control allows various synthetic camera trajectories to render a person. Our experiments show that our method is resilient in generating prolonged motions and varied challenging and complex poses compared to prior methods.
    Please check our website for more details: \href{https://boyiliee.github.io/3DHM.github.io/}{3DHM.github.io}.
% \vspace{-0.5cm}

\end{abstract}

\section{Introduction}

\begin{figure*}[!htb]
    \centering
    \includegraphics[width=\linewidth]{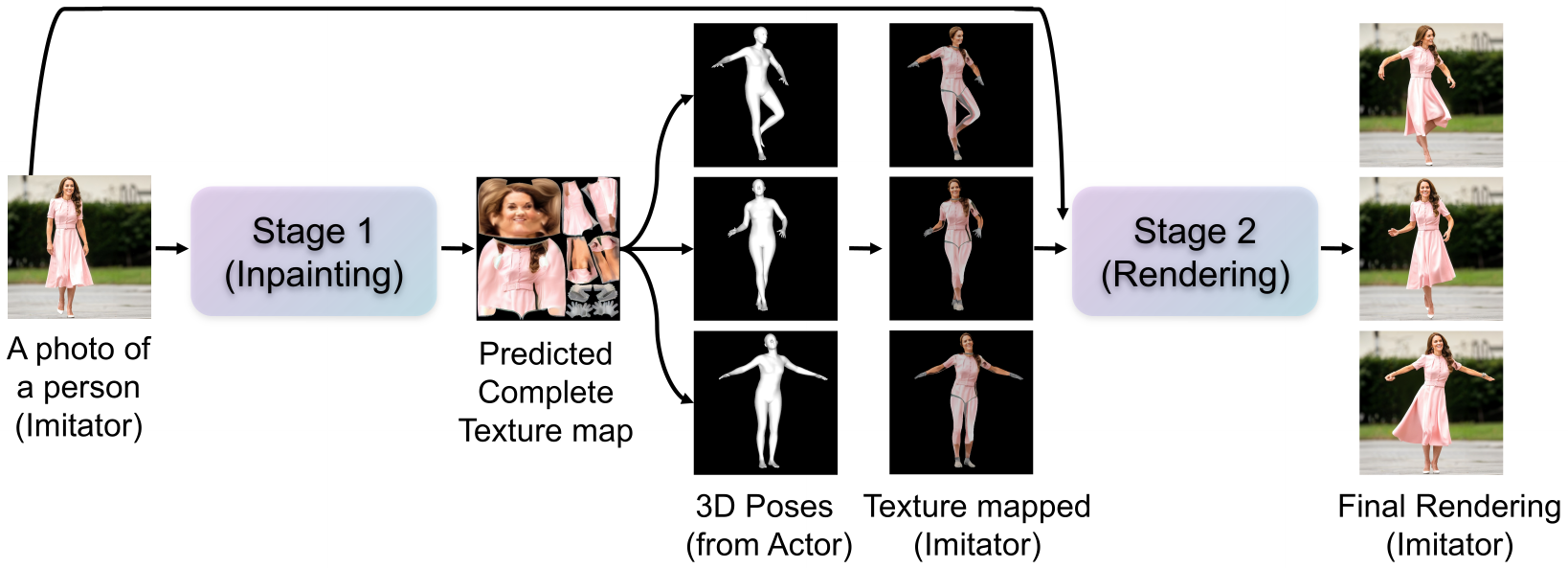}
    \caption{\textbf{Overview of \methodname{}:} we show an overview of our model pipeline. Given an image of the imitator and a sequence of 3D poses from the actor, we first generate a complete full texture map 
    of the imitator, which can be applied to the 3D pose sequences extracted from the actor to generate texture-mapped intermediate renderings of the imitator. Then we pass these intermediate renderings to the Stage-2 model to project the SMPL mesh rendering to more realistic renderings of real images. }
    \label{fig:framework}
\end{figure*}

Given a random photo of a person, can we accurately animate that person to imitate someone else's action?  
This problem requires a deep understanding of how human poses change over time, learning priors about human appearance and clothing. 
For example, in Figure~\ref{fig:teaser} the \textbf{Actor} can do a diverse set of actions, from simple actions such as walking and running to more complex actions such as fighting and dancing. For the \textbf{Imitator}, learning a visual prior about their appearance and clothing is essential to animate them at different poses and viewpoints.
To tackle this problem, we propose \textbf{\methodname{}}, a diffusion framework (see Figure~\ref{fig:framework}) that synthesizes \textbf{3D} \textbf{H}uman \textbf{M}otions by completing a texture map from a single image and then rendering the 3D humans to imitate the actions of the actor.

We use state-of-the-art 3D human pose recovery model 4DHumans~\citep{rajasegaran2022tracking, goel2023humans} for extracting motion signals of the actor, by reconstructing and tracking them over time.
Once we have a motion signal in 3D, as a sequence of meshes, one would think we can simply re-texture them with the texture map of the imitator to get an intermediate rendering of the imitation task. However, this requires a complete texture map of the imitator.
When given only a single view image of the imitator, we see only a part of their body, perhaps the front side, or the backside but never both sides. To get the complete texture map of the imitator from a single view image, we learn a diffusion model to in-fill the unseen regions of the texture map. This essentially learns a prior about human clothing and appearance. For example, a front-view image of a person wearing a blue shirt would usually have the same color at the back.
With this complete texture map, now we can get an intermediate rendering of the imitator doing the actions of the actor. Intermediate rendering means, wrapping the texture map on top of the SMPL~\cite{loper2023smpl} mesh to get a body-tight rendering of the imitator.

However, the SMPL~\cite{loper2023smpl} mesh renderings are body-tight and do not capture deformations on clothing, like skirts or various hairstyles.
To solve this, we learn a second model, that maps from mesh renderings to more realistic images, by controlling the motion with 3D poses. We find out such a simple framework could successfully synthesize realistic and faithful human videos, particularly for long video generations. We show that the 3D control provides a more fine-grained and accurate flow of motion and captures the visual similarities of the imitator faithfully.

While there has been a lot of work on rewriting the motion of an actor~\cite{bregler2023video,karras2023dreampose,wang2023disco}, each requires either large amounts of data, supervised control signals, or requires careful curations of the training data. For example, Make-a-video~\citep{singer2022make} can generate decent results while for human videos, it often generates incomplete or nonconsequential videos and fails at faithful reconstruction of humans. Some works~\citep{chan2019everybody} use Openpose~\citep{cao2017realtime} as intermediate supervision. However, Openpose primarily contains the anatomical key points of humans, it can not be used to indicate the body shape, depth, or other related human body information. DensePose~\citep{guler2018densepose} aims to recover highly accurate dense correspondences between images and the body surface to provide dense human pose estimation. However, it can not reflect the texture information from the original inputs. Compared to this line of work, ours fully utilizes the 3D models to control the motion, by providing an accurate dense 3D flow of the motion, and the texture map representation makes it easy to learn appearance prior from a few thousand samples.

\section{Related Works}

\textbf{Controllable Human Generation.} Human generation is not an easy task. Unlike image translation~\citep{li2022sitta}, generating different humans requires the model to understand the 3D structure of the human body. Given arbitrary text prompts or pose conditions~\citep {brooks2022hallucinating,kulal2023affordance}, we often find out that existing generative models often generate unreasonable human images or videos. Diffusion-HPC~\citep{weng2023diffusion} proposes a diffusion model with Human Pose Correction and finds that injecting human body structure priors within the generation process could improve the quality of generated images. ControlNet~\citep{zhang2023adding} is designed on neural network architecture to control pre-trained large diffusion models to support additional input conditions, such as Openpose~\citep{cao2017realtime}. GestureDiffuCLIP~\citep{ao2023gesturediffuclip} designs a neural network to generate co-speech gestures. However, these techniques are not tailored for animating humans, and do not guarantee the required human appearance and clothing.

\noindent\textbf{Synthesizing Moving People.}
Synthesizing moving people is very challenging. For example, Make-a-Video~\citep{singer2022make} or Imagen Video~\citep{saharia2022photorealistic} could synthesize videos based on a given instruction. However, the generated video cannot accurately capture human properties correctly and may cause the weird composition of generated humans. Older methods~\cite{chan2019everybody,wang2018video} learn pose-to-pixels mapping directly, but they require being trained for each new person separately. Recent works such as SMPLitex~\citep{casas2023smplitex} consider human texture estimation from a single image to animate a person. However, there is a visual gap between rendered people via predicted texture map and real humans. Many works start to directly predict pixels based on diffusion models, such as Dreampose~\citep{karras2023dreampose}, DisCO~\citep{wang2023disco}, AnimateAnyone~\cite{hu2024animate}, MagicAnimate~\cite{xu2024magicanimate}, Champ~\cite{zhu2024champ}, etc.~\cite{chang2023magicdance, wang2024unianimate,ma2024follow, peng2024controlnext}. DreamPose and MagicAnimate is controlled by DensePose~\citep{guler2018densepose}, it aims to synthesize a video containing both human and fabric motion based on a sequence of human body UV or Segmentation maps. DisCO and AnimateAnyone is directly controlled by Openpose~\citep{cao2017realtime}, and it aims to animate the human based on the 2D pose information. Champ~\cite{zhu2024champ} utilizes the multiple condition maps rendered from SMPL mesh to further enhance detailed controllability. However, the approach of aligning output pixels for training regularization often leads these models to become overly specialized to certain training data. Moreover, this methodology limits the models' generalization, as they often perform well on a few people whose data distribution closely matches that of the training dataset.

\section{Synthesizing Moving People}

In this section, we discuss our two-stage approach for imitating a motion sequence. Our ~\methodname{} framework embraces the advantage of accurate 3D pose prediction from the state-of-the-art predicting models 4DHumans~\citep{rajasegaran2022tracking,goel2023humans}, which could accurately track human motions and extracts 3D human poses of the actor videos. For any given video of the actor we want to imitate, we use 3D reconstruction-based tracking algorithms to extract 3D mesh sequences of the actor. For the inpainting and rendering part, we rely on the pre-trained Stable Diffusion~\citep{rombach2022high} model, which is one of the most recent classes of diffusion models that achieve high competitive results over various generative vision tasks. 

Our approach \methodname{} is composed of two core parts: \sonemethod{} for texture map in-painting as Stage-1 and \stwomethod{} for human rendering as Stage-2. Figure~\ref{fig:framework} shows a high-level overview of our framework.
In Stage-1, first, for a given single view image, we extract a rough estimate of the texture map by rendering the meshes onto the image and assigning pixels to each visible mesh triangle such that when rendered again it will produce a similar image as the input image. This predicted texture map has only visible parts of the input image. 
The Stage-1 Diffusion in-painting model takes this partial texture map and generates a complete texture map including the unseen regions. Given this complete texture map, we generate intermediate renderings of SMPL~\cite{loper2023smpl} meshes and use Stage-2 model to project the body-tight renderings to more realistic images with clothing. For the Stage-2 model, we apply 3D control to animate the imitator to copy the actions of the actor.

\subsection{Texture map Inpainting}
\label{sec:stage1}
The goal of Stage-1 model is to produce a plausible complete texture map by inpainting the unseen regions of the imitator. We extract a partially visible texture map by first rendering a 3D mesh onto the input image and sample colors for each visible triangle following 4DHumans~\cite{goel2023humans}.

\noindent\textbf{Input.} We first utilize a common approach to infer pixel-to-surface correspondences to build an incomplete UV texturemap~\citep{xu20213d,casas2023smplitex} for texturing 3D meshes from a single RGB image. We also compute a visibility mask to indicate which pixels are visible in 3D and which ones are not.

\noindent\textbf{Target.} We train our model on a large 3D human texture dataset~\cite{liu2024texdreamer}, which contains 50k high-fidelity textured UV map of SMPL~\cite{loper2023smpl}. To strengthen the model's 3D geometry consistency in completing the partial texturemap, We densely sample a group of visibility masks from 360 degrees of SMPL mesh, which then mask out Ground-Truth texture map to produce the pseudo-partial texture map during training the inpainting model. Benefiting from the extensive collection of texture maps from diverse human appearances, as well as the numerous visibility masks from various viewpoints.

\begin{figure}[t]
    \centering
    \includegraphics[width=\linewidth]{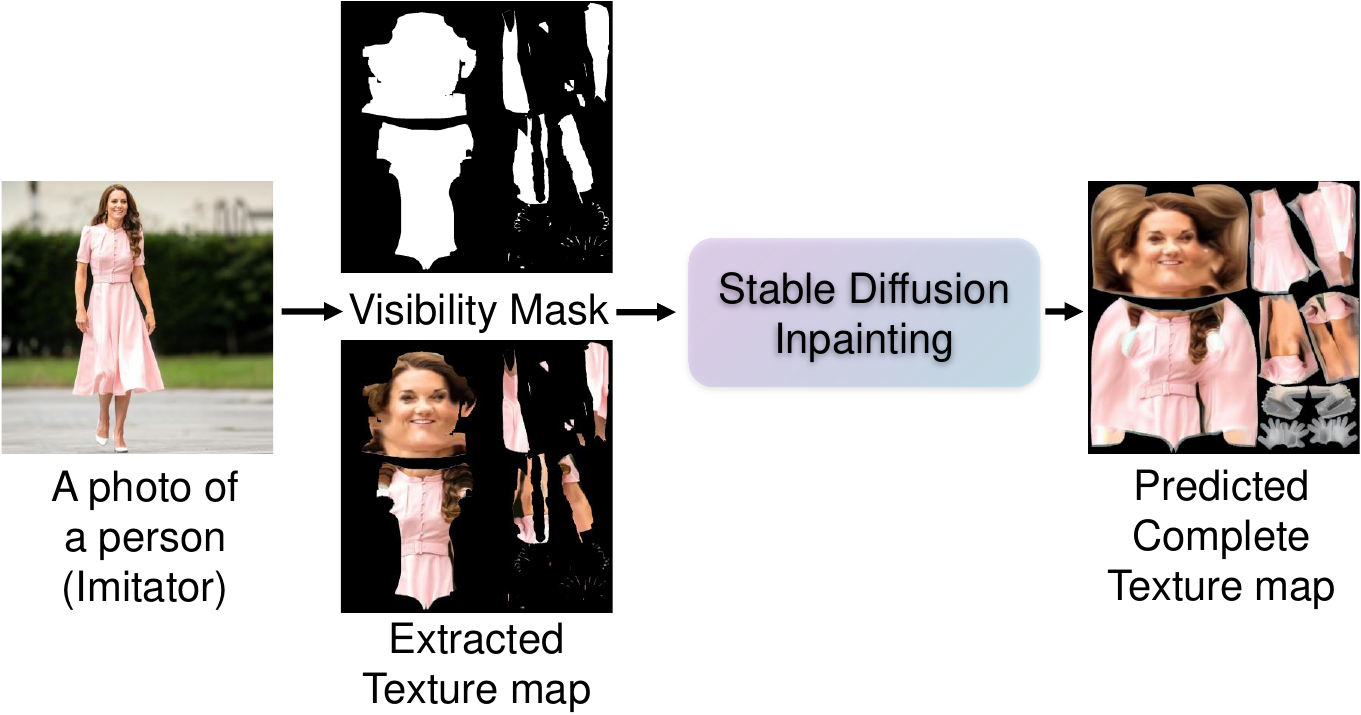}
    \caption{\textbf{Stage-1 of \methodname{}:} In the first stage, given a single view image of an imitator, we first apply 4Dhumans~\cite{goel2023humans} style sampling approach to extract partial texture map and its corresponding visibility map. These two inputs are passed to the in-painting diffusion model to generate a plausible complete texture map. In this example, while we only see the \textbf{front view} of the imitator, the model was able to hallucinate a plausible back region that is consistent with their clothing. }
    \label{fig:s1model}
\end{figure}

\noindent\textbf{Model.} We finetune directly on the Stable Diffusion Inpainting model~\cite{Rombach_2022_CVPR} that shows great performance on image completion tasks. Given a single RGB human image, we predict the human mesh and calculate its corresponding visibility mask and partial texture map, which is then recovered by the in-painting model to complete texture map for the human. We lock the text encoder branch during training and feed "3D realistic human, UV texturemap" as input text condition. We refer to our trained model as \sonemethod{}. See Figure~\ref{fig:s1model} for the model architecture.

\subsection{Human Rendering}
\label{sec:stage2}
In Stage 2, we aim to obtain a realistic rendering of a human imitator doing the actions of the actor.
While the intermediate renderings (rendered with the poses from the actor and texture map from Stage-1) can reflect diverse human motion, these SMPL mesh renderings are body-tight and cannot represent realistic rendering with clothing, hairstyles, and body shapes. 
We train a model for realistic rendering, in a fully self-supervised fashion, by relying on the actor as the imitator. We obtain a sequence of poses from 4DHumans~\cite{goel2023humans} for each training video and use Stage-1 on single frames to obtain a complete texture map. We then pair the intermediate renderings (i.e. the rendered texture maps on the 3D poses) with the original frames from which they were obtained. We collect a large amount of paired data and train our Stage-2 diffusion model with conditioning.

\noindent\textbf{Input:} We first apply the generated complete texture map from Stage-1 to the actor's 3D body mesh sequences to obtain the intermediate rendering. Note that the rendering can only reflect the clothing that fits the 3D mesh (body-tight clothing) but fails to reflect the texture outside the SMPL body (e.g., the puffed-up skirt region, or hat). To obtain the human with complete clothing texture, we input the obtained intermediate renderings and the original image of the person into \stwomethod{} to render the human in a novel pose with a realistic appearance.

\noindent\noindent\textbf{Target:} Since we collected the data by assuming the actor is the imitator, we have the paired data of the intermediate renderings and the real RGB images. This allows us to train this model on lots of data, without requiring any direct 3D supervision.

\begin{figure}[!t]
    \centering
    \includegraphics[width=0.9\linewidth]{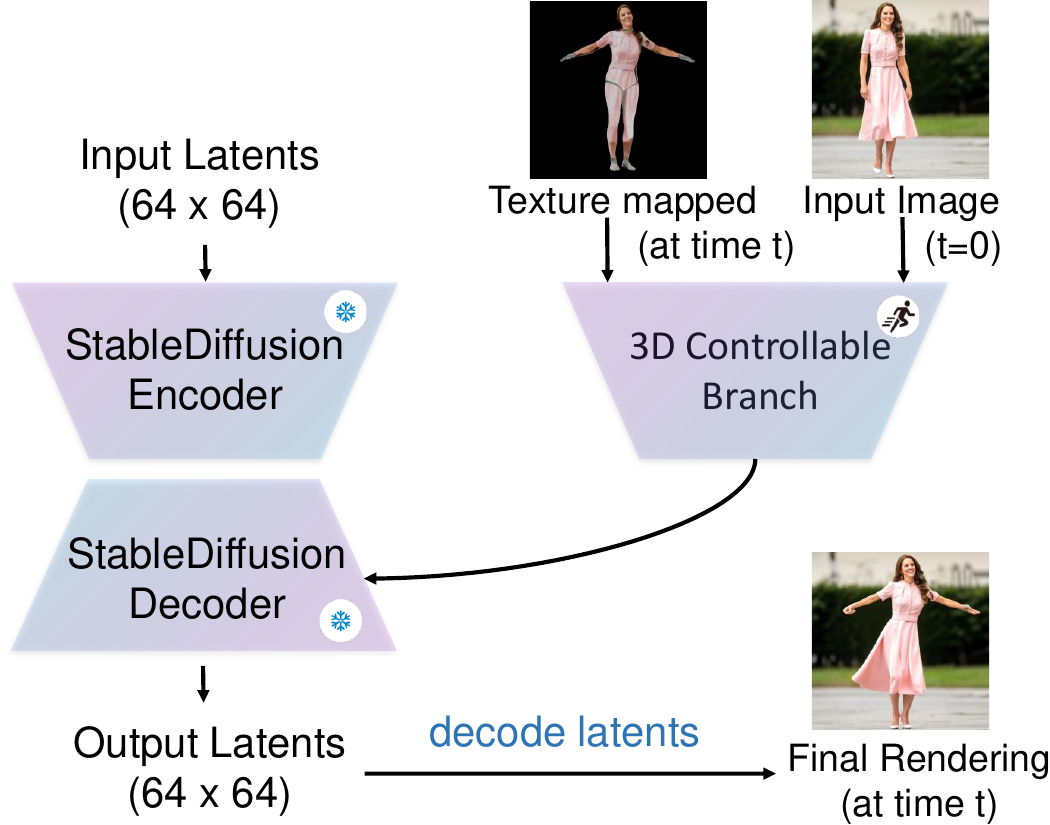}
    \caption{\textbf{Stage-2 of \methodname{}:} Given an intermediate rendering of the imitator with the pose of the actor and the actual RGB image of the imitator, our model can synthesize realistic renderings of the imitator on the pose of the actor.}
    \label{fig:s2model}
    \vspace{-0.2cm}
\end{figure}

\noindent\textbf{Model.} Similar to ControlNet, we directly clone the weights of the encoder of the Stable Diffusion~\cite{rombach2021high} model as our Controllable branch ("trainable copy") to process 3D conditions. We freeze the pre-trained Stable Diffusion.  In the meanwhile, we input a texture-mapped 3D human at time $t$ and original human photo input into a fixed VAE encoder and obtain texture-mapped 3D human latents ($64 \times 64$) and appearance latents ($64 \times 64$) as conditioning latents. We feed these two conditioning latents into \stwomethod{} Controllable branch. The key design principle of this branch is to learn textures from human input and apply them to the texture-mapped 3D human during training through the denoising process. The goal is to render a real human with vivid textures from the generated(texture-mapped) 3D human from Stage 1. We obtain the output latent and process it to the pixel space via diffusion step procedure and fixed VAE decoder. 
We refer to our trained model as \stwomethod{}. In \stwomethod{}, we predict outputs frame by frame. We show the Stage 2 workflow in Figure~\ref{fig:s2model}.

\section{A Complete Approach}

\begin{figure}[t]
    \centering
    \includegraphics[width=\linewidth]{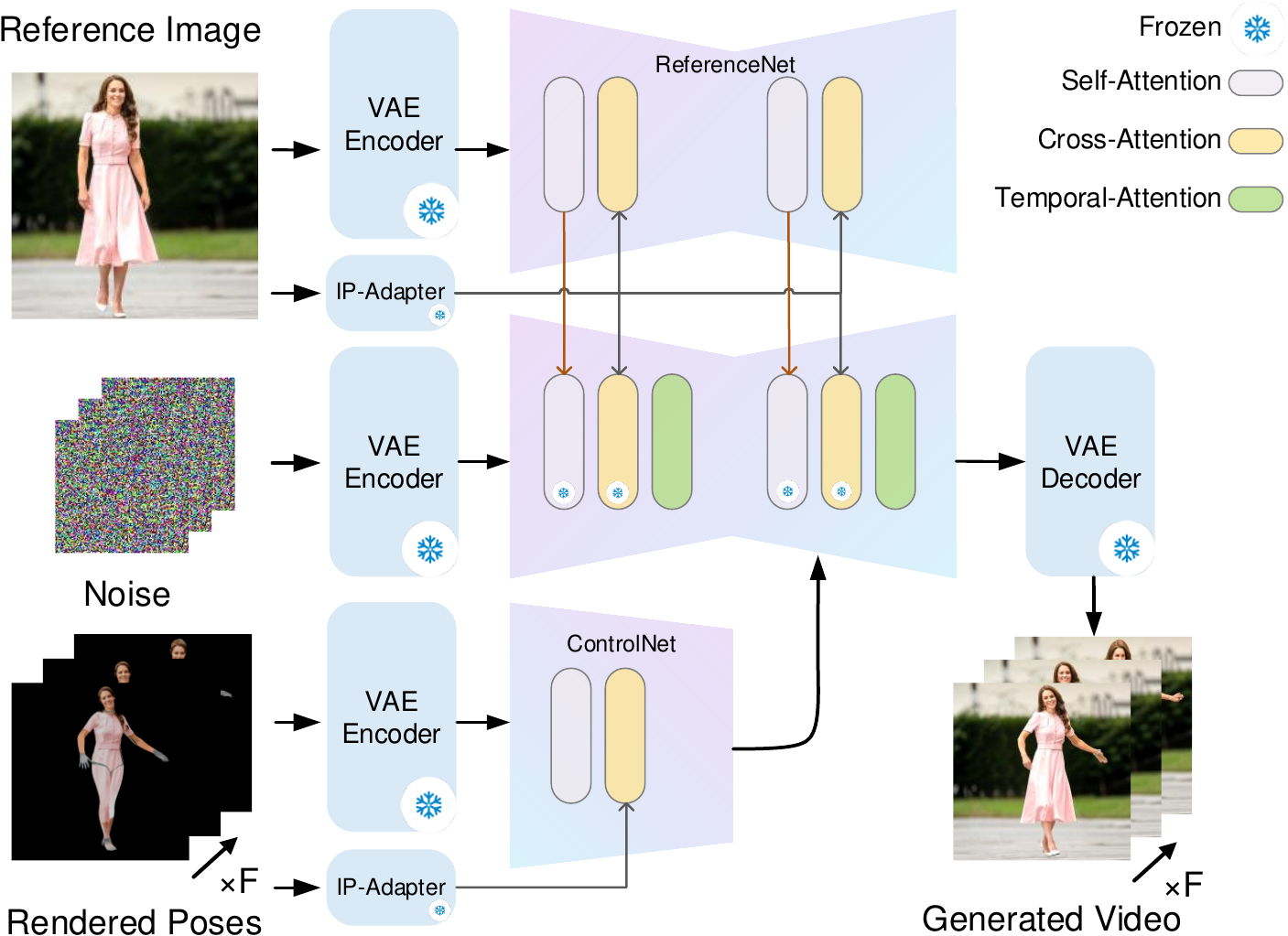}
    \vspace{-0.1in}
    \caption{\textbf{Scaled up Stage-2 of \methodname{} Model:} To enable consistent background and human generation, we train ReferenceNet with ControlNet, and then only finetune the temporal-attention layer of the UNet and keep other components frozen.}
    \vspace{-0.2in}
    \label{fig:realmodel}
\end{figure}

In this section, we discuss how to scale up our method to real-world domains. We first discuss the challenges of visual appearance injection for Diffusion models and then propose a novel framework to enhance consistent appearance alignment. For any given reference human images (the imitators), our network can generate high-fidelity results benefiting from the enhanced appearance encoder~\citep{ye2023ip}. To ensure the visual consistency of both human identity and background from reference images, we use a trainable copy of base Stable Diffusion model~\citep{rombach2022high} to inject appearance information that perfectly aligns with the backbone denoising Diffusion model. To encourage temporally smooth reconstructions, we utilize a temporal diffusion model~\citep{guo2023animatediff} to learn the temporal correlation within motion sequences. To generate smooth and consistent videos, we propose an easy but efficient method that takes the previous frame for consequential video generation. The detailed framework is shown in Figure~\ref{fig:realmodel}.

\subsection{Enhance Appearance Alignment}
\label{sec:appearance}
The key challenge in scaling up our method to real-world domains is to both maintain the complex background and the human appearance from reference images consistently. The Stable Diffusion Model~\cite{rombach2022high} is trained on text-to-image tasks, and focuses on semantics of the generated image. However, in our Stage-2 rendering task, instead of semantic features, the model needs more low-level visual features to reconstruct the detailed structure and appearance from the input images. Given the imitator's images as reference, our approach simultaneously leverages the capabilities of Stable Diffusion and uses the reference image prompts to obtain more accurate generation. We use a lightweight image adapter~\cite{ye2023ip} to condition diffusion on image prompts. We also add a trainable branch of the Stable Diffusion model, namely Reference Net, to enhance appearance consistency on both the input image's background and human appearance.

\noindent\textbf{Input.} 
Same with Stage-2, we input the generated imitator's texture map from Stage-1 with actor's 3D mesh sequences to get the intermediate rendering. Then the intermediate rendering is input to the 3D Controllable branch as motion condition. The original imitator's RGB image is input to the Reference Net and the image adaptor as the appearance guidance.

\noindent\textbf{Target.} 
We scale up our model on about 1000 real human videos collected from the Internet, each containing 2-10 seconds solo dancing. For this stage, we keep the assumption that: imitator and actor are identical and randomly sample the imitator and actor. The objective now is to drive the reference imitator image with actor's pose to generate the corresponding target image. We trained on our real and virtual datasets together to teach model focus on complex pose and appearance variance, and the 3D view consistency respectively.

% \vspace{0.2cm}

\noindent\textbf{Model.} To better align the input image's appearance with the denoising backbone, we make a trainable copy of the pre-trained Stable Diffusion as our ReferenceNet. Notice that different from the Section 3.2, instead of adding appearance latents with 3D human latents together, we now separately input the imitator's appearance latents to ReferenceNet to extract level of details appearance latents. The appearance latents are then injected into the denoising backbone to condition the denoising process, which help to keep consistent background and human appearance for different poses. The pretrained IP-adapter~\cite{ye2023ip} is also integrated into all branches to keep human identity.

\subsection{Temporal Consistency}
\label{sec:temporal}
Once our image-to-image model learned to generate the imitator frames we can apply it frame-by-frame for image-to-video generation. However, the generated video may suffer from jittering due to the lack of temporal consistency. Based on the Stage-2 model, we adapt a temporal model pretrained on a large video dataset ~\cite{guo2023animatediff} to learn the motion's coherence and appearance consistency. Though previous works ~\cite{wang2023disco, karras2023dreampose, hu2024animate} also have similar temporal layers and success for short video, they still suffer from inconsistency in long video generation results. To effectively solve the unstable and randomness between each short video clip generated from temporal sliding windows, we design an easy process to concatenate the previous frame's latents with the consequential video.

\noindent\textbf{Input.} In this stage, our model is optimized directly on video data. During training, for each video clip, we extract $F$ consecutive frames as the target of actor's motion sequence and randomly pick a frame as the imitator's reference image. Now the 3D Control branch takes $F$ consecutive intermediate rendering as motion sequences to drive the imitator's image to generate a temporal coherent video. Notice that our Reference Net here will not cost extra computing time since there is only one reference image for each video clip.

\noindent\textbf{Model.} 
The short video clip is now defined as $V\in\mathbb{R}^{B \times C \times F \times H \times W}$, with batch size $B$, the number of channels $C$, the number of consecutive frames $F$, height $H$ and width $W$ respectively. The temporal layers are inserted at each resolution level. For level $i$ , the 5D latents $v_{i}\in\mathbb{R}^{b \times c \times f \times h \times w}$ is reshaped to ${(b \times h \times w) \times f \times c}$ within the temporal layer for self-attention to align feature maps across frames. The temporal attention mechanism above not only effectively smooths the flickering and jittering, but also improves the motion and appearance consistency in generated videos. However, during long video generation, since the video clips are generated independently and concatenated together, the different random noises will cause inconsistency between each video clip. To facilitate the cross-clips consistency, we take the last frame $V_{k}^{f}$ from $k$-th generated video clip $V_{k}^{1:f}$ to condition on the next video clip generation $V_{k+1}^{1:f}$. The $V_{k}^{f}$ is input to the Reference Net to extract corresponding latents for each resolution level $i$, and then fed into the temporal layers concatenated with original 5D latents along the temporal dimension. The conditioned temporal layers at level $i$ now attention across latent $v_{i}^{0:f}\in\mathbb{R}^{b \times c \times (1+f) \times h \times w}$ and then trunck the previous frame $v_{i}^{0}$ to get the conditioned results $\bar{v}_{i}^{1:f}$. During this stage, we freeze all other parameters and train only the temporal model. Zero-initialize~\cite{zhang2023adding} is also applied to the temporal layers to eliminate harmful noise during training. 

\section{Experiments}

\subsection{Experimental Setup}
\textbf{Dataset.} We collect 2,524 3D human videos from 2K2K~\citep{han2023high}, THuman2.0~\citep{tao2021function4d} and People-Snapshot~\citep{alldieck2018video} datasets. 2K2K is a large-scale human dataset with 3D human models reconstructed from 2K resolution images. THuman2.0 contains 500 high-quality human scans captured by a dense DLSR rig. People-Snapshot is a smaller human dataset that captures 24 sequences. We convert the 3D human dataset into videos and extract 3D poses from human videos using 4DHumans. We use 2,205 videos for training and other videos for validation and testing. See the Appendix for more details on the dataset distribution on clothing.

\noindent\textbf{Evaluation Metrics.} We evaluate the quality of generated frames of our method with image-based and video-based metrics.  
For image-based evaluation, we follow the evaluation protocol of DisCO~\citep{wang2023disco} to evaluate the generation quality. We report the average PSNR~\citep{hore2010image}, SSIM~\citep{wang2004image}, FID~\citep{heusel2017gans}, LPIPS~\cite{zhang2018unreasonable}, and L1. For video-based evaluation, we use FVD~\citep{unterthiner2018towards}. For pose evaluating 3D pose accuracy we use Mean Per-Vertex Position Error (MPVPE) and Procrustes-Aligned Mean Per-Vertex Position Error (PA-MVPVE~\citep{moon2022accurate}).

\noindent\textbf{Implementation Details.} We set a learning rate of 5e-05 and use the pre-trained diffusion models for both stages. For Stage 1 \sonemethod{}, we fine-tune Stable Diffusion Inpainting models~\citep{Rombach_2022_CVPR}
We train \stwomethod{} for 50 epochs (requires about 2 weeks on our compute). For Stage 2 \stwomethod{}, we train the Controllable branch and freeze Stable Diffusion backbones. The total number of trainable parameters in this case is 876M. 
We train \stwomethod{} for 30 epochs (requires about 2 weeks on 8 NVIDIA A100 GPUs with a batch size of 4). During inference, we only need to run Stage-1 once to reconstruct the full texture map of the imitator, which is used for all other novel poses and viewpoints. We run Stage-2 inference for each frame independently, however since the initial RGB frame of the imitator is conditioned for all frames, the Stage-2 model is able to produce samples that are temporarily consistent.

\begin{figure*}[!t]
\centering
\begin{subfigure}[b]{1\textwidth}
\centering
   \includegraphics[width=0.9\linewidth]{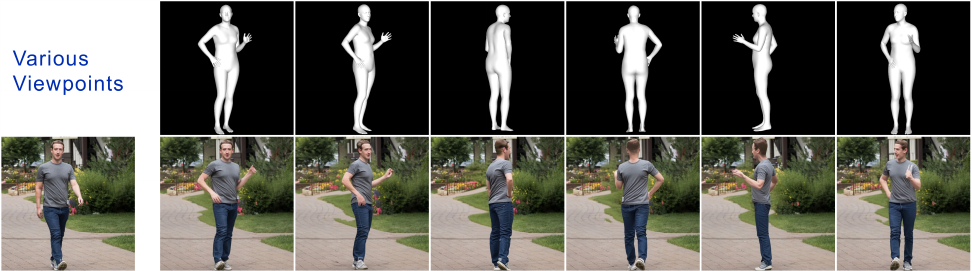}
   \caption{3DHM with random 3D poses from various viewpoints. We show that even if the person's photo is from a side angle, our stage 1 can help reconstruct the full texture map, which could be used to obtain full body information. Stage 2 can add texture information based on a given input.  }
   \label{fig:showresult3D}
\end{subfigure}
\begin{subfigure}[b]{1\textwidth}
\centering
   \includegraphics[width=0.9\linewidth]{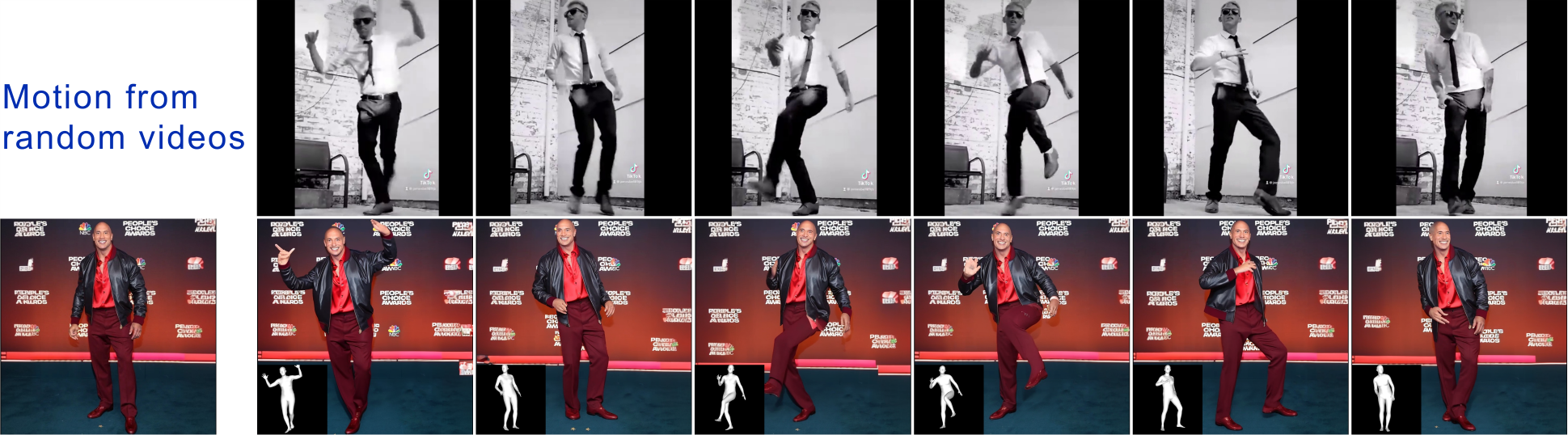}

   \caption{3DHM with motions from random YouTube Videos.}
   \label{fig:showresultrand}
\end{subfigure}
\caption{Qualitative results on different viewpoints of the same pose and motions from random videos. }
\vspace{-0.1in}
\label{fig:showall}
\end{figure*}

\subsection{Quantitative Results}
\noindent\textbf{Baselines.}
We compare our approaches with past and state-of-the-art methods: DreamPose~\citep{karras2023dreampose}, DisCo~\citep{wang2023disco} and ControlNet~\citep{zhang2023adding} (for pose accuracy comparisons)\footnote{We utilize the open-source official code and models provided by the authors to implement these baselines. We use diffusers~\citep{diffusers-von-platen-etal-2022} for ControlNet and Openpose extraction, and Detectron2 for DensePose extraction for MagicAnimate and DreamPose. Since ~\citet{chan2019everybody} can only work for animating a specific person, we don't compare with it in this paper.}. We set inference steps as 50 for all the approaches for fair comparisons.

\noindent\textbf{Comparisons on Frame-wise Generation Quality.}
We compare ~\methodname{} with other methods on 2K2K test dataset, which is composed of 50 unseen human videos, at $256 \times 256$ resolution. For each human video, we take 30 frames that represent the different viewpoints of each unseen person. The angles range from $0^{\circ}$ to $360^{\circ}$, we take one frame every $12^{\circ}$ to better evaluate the prediction and generalization ability of each model. As for DisCO, we strictly follow their setting and extract OpenPose for inference. We extract DensePose for inference DreamPose and MagicAnimate. We evaluate the results and calculate the average score over all frames of each video. We set the background as black for all approaches for fair comparisons. We report the average score of the same 50 videos and show the comparisons in Table~\ref{tab:compare_generation}. We observe that \methodname{} outperforms all the baselines in different metrics.
\begin{table}[t]
    \centering
    {
    \resizebox{1\linewidth}{!}{
    \begin{tabular}{c|c|c|c|c|c|c|c}
    \toprule 
    Method & PSNR$\uparrow$&SSIM $\uparrow$& FID $\downarrow$ & LPIPS $\downarrow$& L1 $\downarrow$ & FID-VID$\downarrow$& FVD $\downarrow$\\
    \midrule\midrule
    DreamPose& 35.06&	0.80 &	245.19&0.18&2.12e-04&113.96 &950.40 \\
    DisCO& 35.38& 0.81&164.34&0.15& 1.44e-04&83.91 & 629.18 \\
    MagicAnimate&32.57 &0.65&300.66&0.29&5.80E-04&140.45&900.70 \\
    \midrule
    \rowcolor{aliceblue}
    Ours  & \textbf{36.18}&\textbf{0.86}&\textbf{154.75}&\textbf{0.12}&\textbf{9.88e-05}&\textbf{55.40}&\textbf{422.38}\\
    \bottomrule
    \end{tabular}
    }
    }
    \vspace{-0.1in}
    \caption{\textbf{Quantitative comparison on generation quality:} We compare our method with prior works on pose condition generation tasks and measure the generation quality of the samples.}
    \label{tab:compare_generation}
    \vspace{-0.2in}
\end{table}

\noindent\textbf{Comparisons on Video-level Generation Quality.}
To verify the temporal consistency of \methodname{}, we also report the results following the same test set and baseline implementation as in image-level evaluation. Unlike image-level comparisons, we concatenate every consecutive 16 frames to form a sample of each unseen person on challenging viewpoints. The angles range from $150^{\circ}$ to $195^{\circ}$, we take one frame every $3^{\circ}$ to better evaluate the prediction and generalization ability of each model. We report the average score overall of 50 videos and show the comparisons in Table~\ref{tab:compare_generation}. We observe that \methodname{}, though trained and tested by per frame, still embraces significant advantage over prior approaches, indicating superior performance on preserving the temporal consistency with 3D control.

\noindent\textbf{Comparisons on Pose Accuracy.}
To further evaluate the validity of our model, we estimate 3D poses from generated human videos from different approaches via a state-of-the-art 3D pose estimation model 4DHumans. We use the same dataset setting mentioned above and compare the extracted poses with 3D poses from the target videos. Following the same comparison settings with generation quality, we evaluate the results and calculate the average score over all frames of each video. Beyond DreamPose and DisCO, we also compare with ControlNet, which achieves the state-of-the-art in generating images with conditions, including openpose control. Since ControlNet does not input images, we input the same prompts as ours `a real human is acting' and the corresponding openpose as conditions. 
We report the average score overall of 50 test videos and show the comparisons in Table~\ref{tab:compare_pose}. We could notice that \methodname{} could synthesize moving people following the provided 3D poses with very high accuracy. At the same time, previous approaches might not achieve the same performance by directly predicting the pose-to-pixel mapping. We also notice that \methodname{} could achieve better results on both 2D metrics and 3D metrics, even if DisCO and ControlNet are controlled by Openpose and DreamPose is controlled by DensePose. 
\begin{table}[t]
    \centering
    {
    \begin{tabular}{c|c|c}
    \toprule 
    Method & MPVPE $\downarrow$ & PA-MPVPE $\downarrow$ \\
    \midrule\midrule
    DreamPose  	  &	 123.07 & 82.75 \\
    DisCO         &  112.12 & 63.33 \\ 
    ControlNet    &  108.32 & 59.80 \\ \midrule
    \rowcolor{aliceblue}
    Ours  	     &  \textbf{41.08} & \textbf{31.86} \\
    \bottomrule
    \end{tabular}
    }
    \vspace{-0.1in}
    \caption{Quantitative comparison on pose accuracy. We measure the pose accuracy in the generated images given the conditioned pose as the ground truth. We can see that our model is very accurate in persevering the poses in generated images.}
    \label{tab:compare_pose}
    \vspace{-0.2in}
\end{table}

\subsection{Ablation Study}
To further verify the components of our methods, we train on training dataset and test on test datasets. We extract the 3D rendered pose from these 50 test video tracks. Same with the settings in quantitative comparison, we calculate the average scores among all the generated frames and targeted original frames and report the results on both frame-wise metric (PSNR, SSIM, FID, LPIPS, L1), video-level metric (FID-VID, FVD) and pose accuracy (MPVPE, PA-MPVPE) in Table \ref{tab:compare_components}. We find that both texture map reconstruction and appearance latents are critical to the model performance. 
Also, we notice that directly adding SMPL parameters into the model during training may not bring improved performance considering all evaluation metrics. 

\noindent\textbf{Running Cost.}
Here we outline the comparison of parameters and running time with other methods in Table~\ref{tab:compare_generation} using a single GPU A100. We show the comparison in Table~\ref{tab:runningcost}.
\begin{table}[]
\centering
    {
    \resizebox{0.8\linewidth}{!}{
    \begin{tabular}{c|c|c}
    \toprule 
    Method &Time (second/frame) & Parameter \\
    \midrule\midrule
    DreamPose&22.0 &1.0B \\
    DisCO&5.0 &2.0B \\
    MagicAnimate&10.0 &2.0B \\ \hline
    \rowcolor{aliceblue}
    Ours&\textbf{3.2} &2.0B \\
    \bottomrule
    \end{tabular}
    }
    }
    \vspace{-0.1in}
    \caption{Comparison of running cost. We compare inference time for different models, and we can see that our models is faster in comparison with our models.}
    \vspace{-0.3cm}
    \label{tab:runningcost}
\end{table}

\section{Analysis and Discussion}

\begin{table*}[!t]
    \centering
    {
    \resizebox{1\linewidth}{!}{
    \begin{tabular}{c|c|c|c|c|c|c|c|c|c}
    \toprule 
    Settings & PSNR$\uparrow$&SSIM $\uparrow$& FID $\downarrow$ & LPIPS $\downarrow$& L1 $\downarrow$ & FID-VID$\downarrow$& FVD $\downarrow$ &MPVPE $\downarrow$ & PA-MPVPE $\downarrow$\\
    \midrule\midrule
    Default& \underline{36.18}& \underline{0.86}&\textbf{154.75} &\textbf{0.12}&\underline{9.88e-05}&\textbf{55.40} & \textbf{422.38}&  \underline{41.08} & \underline{31.86}\\
    \midrule
    w/o Texture map &35.00 & \textbf{0.78}&237.42&0.20&2.35e-04&113.97 &632.67&  92.94 & 59.18\\
    w/o Appearance Latents &36.07 & \underline{0.86}&167.58&\textbf{0.12}&1.03e-04& 93.21&715.51 &  41.99 & 32.82\\
    adding SMPL parameters & \textbf{36.42}&0.87&\underline{157.60}&\textbf{0.12}& \textbf{8.87e-05}&\underline{72.35} &\underline{579.90}&  \textbf{39.16} & \textbf{29.67}\\
    \bottomrule
    \end{tabular}
    }
    }
    \vspace{-.1in}
    \caption{Ablation study of \stwomethod{}. We compare the frame-wise generation quality, video-level generation quality and the pose accuracy under different settings. We notice both texturemap reconstruction and appearance latents are critical to the model performance. The results show that although adding SMPL parameters achieve better performance on frame-wise setting but may yield worse temporal consistency than default settings. Note: we use \textbf{bold} to represent the best result and \underline{underline} to represent the second-best result.
}
    \vspace{-0.1in}
    \label{tab:compare_components}
\end{table*}

\subsection{Qualitative Results}
Our work focuses on synthesizing moving people, primarily for clothing and the human body. With the aid of 3D assistance, our approach has the potential to produce human motion videos in various scenarios. We consider challenging 3D poses and motions from 2 sources: 3D human videos and random YouTube videos. We utilize our model, which has been scaled up for real-world domains.

\noindent\textbf{Poses from Unseen 3D Human Videos.} 
We test our model on different 3D human videos with different human appearances and 3D poses from the 2K2K dataset. We verify that the tested video has never appeared in training data. We display the results in Figure~\ref{fig:showresult3D}.

\noindent\textbf{Motions from Random YouTube Videos.} 
We test our model on very different motions from randomly downloaded YouTube videos for an unseen human. We display the results in Figure~\ref{fig:showresultrand}. The results show that 3DHM can efficiently animate any person using random motion resources, accurately following the 3D poses from challenging motion sources.

\subsection{Qualitative Comparison}
\begin{figure}[!htbp]
    \centering
    \includegraphics[width=1\linewidth]{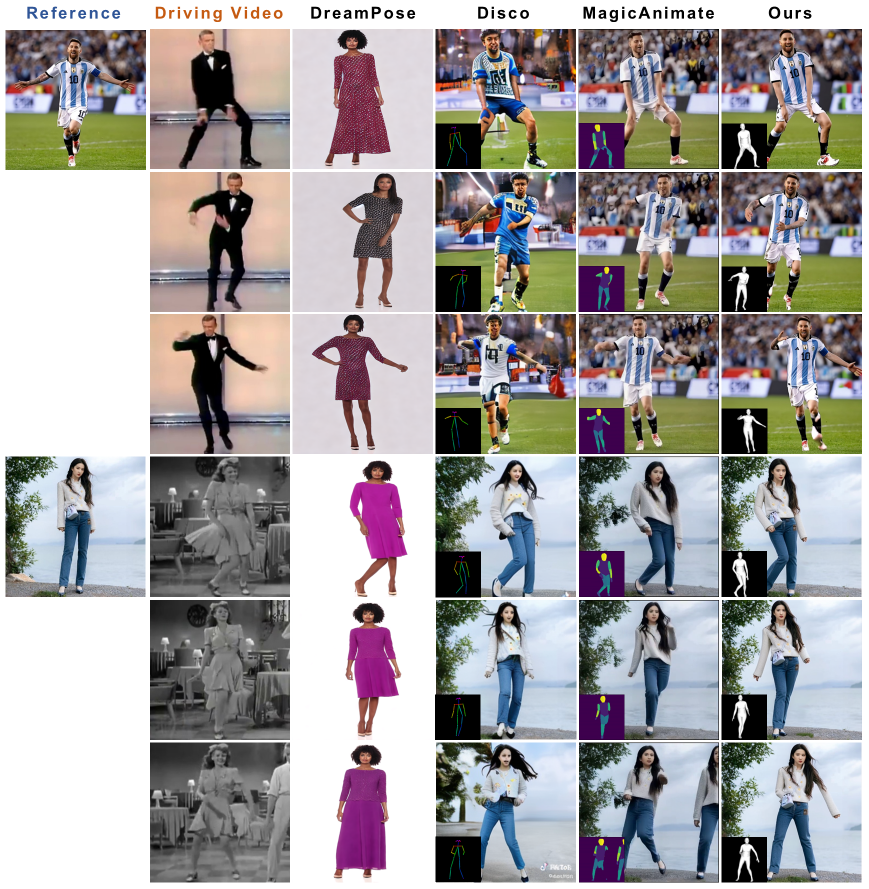}
    \vspace{-0.3in}
    \caption{Qualitative comparison with other state-of-the-art approaches on a real human photo.}
    \label{fig:compare2d3d}
    \vspace{-0.1in}
\end{figure}

We also compare the results of the official model from DreamPose, DisCO, and MagicAnimate on a random person on a random real human photo which ensures distinct data distribution. We display the qualitative results of various poses on real human photos in Figure~\ref{fig:compare2d3d}. We notice that \methodname{} can generalize well to unseen real humans though it is only trained by limited 3D humans. Since DreamPose requires subject-specific finetuning of the UNet to achieve better results, it cannot directly generalize well on a random human photo. As for DisCO, though it has been trained with an effective human attribute pre-training on multiple public datasets for better generalizability to unseen humans, still fails to synthesize people without the target pose. MagicAnimate uses 3D pose features (DensePose) which better controls the appearance of input images. However it always suffers from severe artifacts on DensePose segmentation maps, which severely ruins the pose accuracy and consistency. We assume this is because \methodname{} adds rigid 3D control to better correlate the appearance to the poses, and preserve the body shape. Training with OpenPose or DensePose cannot guarantee the mapping between textures and poses, which makes it hard for the models to generalize.

\subsection{Limitations}
As \methodname{} has been trained with limited data (around 2K synthetic humans and 1K real humans), it might struggle to predict the texture details of the unseen side of the input human photo. However, we believe this issue can be mitigated by scaling up with more human data.

\section{Conclusion}
In this paper, we propose \methodname{}, a two-stage diffusion model-based framework that enables synthesizing moving people based on one random photo and target sequence of human poses. A notable aspect of our approach is that we employ a cutting-edge 3D pose estimation model to generate human motion data, allowing our model to be trained on arbitrary videos without necessitating ground truth labels. Our method is suitable for long-range motion generation, and can deal with arbitrary poses with superior performance over previous approaches, by preserving the poses of the target motion, and clothing, face identities and smoother motion between frames.

\section*{Acknowledgement}
We thank the Machine Common Sense project and ONR MURI award number N00014-21-1-2801. We also thank Google's TPU Research Cloud (TRC) for providing cloud TPUs. We thank Georgios Pavlakos, Shubham Goel, Jane Wu and Karttikeya Mangalam for the constructive feedback and helpful discussions.

\bibliography{reference}

\begin{thebibliography}{43}
\providecommand{\natexlab}[1]{#1}
\providecommand{\url}[1]{\texttt{#1}}
\expandafter\ifx\csname urlstyle\endcsname\relax
  \providecommand{\doi}[1]{doi: #1}\else
  \providecommand{\doi}{doi: \begingroup \urlstyle{rm}\Url}\fi

\bibitem[Alldieck et~al.(2018)Alldieck, Magnor, Xu, Theobalt, and Pons-Moll]{alldieck2018video}
Thiemo Alldieck, Marcus Magnor, Weipeng Xu, Christian Theobalt, and Gerard Pons-Moll.
\newblock Video based reconstruction of 3d people models.
\newblock In \emph{Proceedings of the IEEE Conference on Computer Vision and Pattern Recognition}, pages 8387--8397, 2018.

\bibitem[Ao et~al.(2023)Ao, Zhang, and Liu]{ao2023gesturediffuclip}
Tenglong Ao, Zeyi Zhang, and Libin Liu.
\newblock Gesturediffuclip: Gesture diffusion model with clip latents.
\newblock \emph{arXiv preprint arXiv:2303.14613}, 2023.

\bibitem[Bregler et~al.(2023)Bregler, Covell, and Slaney]{bregler2023video}
Christoph Bregler, Michele Covell, and Malcolm Slaney.
\newblock Video rewrite: Driving visual speech with audio.
\newblock In \emph{Seminal Graphics Papers: Pushing the Boundaries, Volume 2}, pages 715--722. 2023.

\bibitem[Brooks and Efros(2022)]{brooks2022hallucinating}
Tim Brooks and Alexei~A Efros.
\newblock Hallucinating pose-compatible scenes.
\newblock In \emph{European Conference on Computer Vision}, 2022.

\bibitem[Cao et~al.(2017)Cao, Simon, Wei, and Sheikh]{cao2017realtime}
Zhe Cao, Tomas Simon, Shih-En Wei, and Yaser Sheikh.
\newblock Realtime multi-person 2d pose estimation using part affinity fields.
\newblock In \emph{Proceedings of the IEEE conference on computer vision and pattern recognition}, pages 7291--7299, 2017.

\bibitem[Casas and Trinidad(2023)]{casas2023smplitex}
Dan Casas and Marc~Comino Trinidad.
\newblock Smplitex: A generative model and dataset for 3d human texture estimation from single image.
\newblock \emph{arXiv preprint arXiv:2309.01855}, 2023.

\bibitem[Chan et~al.(2019)Chan, Ginosar, Zhou, and Efros]{chan2019everybody}
Caroline Chan, Shiry Ginosar, Tinghui Zhou, and Alexei~A Efros.
\newblock Everybody dance now.
\newblock In \emph{Proceedings of the IEEE/CVF international conference on computer vision}, pages 5933--5942, 2019.

\bibitem[Chang et~al.(2023)Chang, Shi, Gao, Fu, Xu, Song, Yan, Yang, and Soleymani]{chang2023magicdance}
Di Chang, Yichun Shi, Quankai Gao, Jessica Fu, Hongyi Xu, Guoxian Song, Qing Yan, Xiao Yang, and Mohammad Soleymani.
\newblock Magicdance: Realistic human dance video generation with motions \& facial expressions transfer.
\newblock \emph{arXiv preprint arXiv:2311.12052}, 2023.

\bibitem[Goel et~al.(2023)Goel, Pavlakos, Rajasegaran, Kanazawa, and Malik]{goel2023humans}
Shubham Goel, Georgios Pavlakos, Jathushan Rajasegaran, Angjoo Kanazawa, and Jitendra Malik.
\newblock Humans in 4{D}: Reconstructing and tracking humans with transformers.
\newblock In \emph{ICCV}, 2023.

\bibitem[G{\"u}ler et~al.(2018)G{\"u}ler, Neverova, and Kokkinos]{guler2018densepose}
R{\i}za~Alp G{\"u}ler, Natalia Neverova, and Iasonas Kokkinos.
\newblock Densepose: Dense human pose estimation in the wild.
\newblock In \emph{Proceedings of the IEEE conference on computer vision and pattern recognition}, pages 7297--7306, 2018.

\bibitem[Guo et~al.(2024)Guo, Yang, Rao, Liang, Wang, Qiao, Agrawala, Lin, and Dai]{guo2023animatediff}
Yuwei Guo, Ceyuan Yang, Anyi Rao, Zhengyang Liang, Yaohui Wang, Yu Qiao, Maneesh Agrawala, Dahua Lin, and Bo Dai.
\newblock Animatediff: Animate your personalized text-to-image diffusion models without specific tuning.
\newblock \emph{International Conference on Learning Representations}, 2024.

\bibitem[Han et~al.(2023)Han, Park, Yoon, Kang, Park, and Jeon]{han2023high}
Sang-Hun Han, Min-Gyu Park, Ju~Hong Yoon, Ju-Mi Kang, Young-Jae Park, and Hae-Gon Jeon.
\newblock High-fidelity 3d human digitization from single 2k resolution images.
\newblock In \emph{Proceedings of the IEEE/CVF Conference on Computer Vision and Pattern Recognition (CVPR)}, 2023.

\bibitem[Heusel et~al.(2017)Heusel, Ramsauer, Unterthiner, Nessler, and Hochreiter]{heusel2017gans}
Martin Heusel, Hubert Ramsauer, Thomas Unterthiner, Bernhard Nessler, and Sepp Hochreiter.
\newblock Gans trained by a two time-scale update rule converge to a local nash equilibrium.
\newblock \emph{Advances in neural information processing systems}, 30, 2017.

\bibitem[Hore and Ziou(2010)]{hore2010image}
Alain Hore and Djemel Ziou.
\newblock Image quality metrics: Psnr vs. ssim.
\newblock In \emph{2010 20th international conference on pattern recognition}, pages 2366--2369. IEEE, 2010.

\bibitem[Hu(2024)]{hu2024animate}
Li Hu.
\newblock Animate anyone: Consistent and controllable image-to-video synthesis for character animation.
\newblock In \emph{Proceedings of the IEEE/CVF Conference on Computer Vision and Pattern Recognition}, pages 8153--8163, 2024.

\bibitem[Karras et~al.(2023)Karras, Holynski, Wang, and Kemelmacher-Shlizerman]{karras2023dreampose}
Johanna Karras, Aleksander Holynski, Ting-Chun Wang, and Ira Kemelmacher-Shlizerman.
\newblock Dreampose: Fashion image-to-video synthesis via stable diffusion.
\newblock \emph{arXiv preprint arXiv:2304.06025}, 2023.

\bibitem[Kulal et~al.(2023)Kulal, Brooks, Aiken, Wu, Yang, Lu, Efros, and Singh]{kulal2023affordance}
Sumith Kulal, Tim Brooks, Alex Aiken, Jiajun Wu, Jimei Yang, Jingwan Lu, Alexei~A. Efros, and Krishna~Kumar Singh.
\newblock Putting people in their place: Affordance-aware human insertion into scenes.
\newblock In \emph{Proceedings of the IEEE Conference on Computer Vision and Pattern Recognition (CVPR)}, 2023.

\bibitem[Li et~al.(2022)Li, Cui, Lin, and Belongie]{li2022sitta}
Boyi Li, Yin Cui, Tsung-Yi Lin, and Serge Belongie.
\newblock Sitta: Single image texture translation for data augmentation.
\newblock In \emph{European Conference on Computer Vision}, pages 3--20. Springer, 2022.

\bibitem[Liu et~al.(2024)Liu, Zhu, Tang, Zhang, Zhang, Cao, Wang, Wu, and Huang]{liu2024texdreamer}
Yufei Liu, Junwei Zhu, Junshu Tang, Shijie Zhang, Jiangning Zhang, Weijian Cao, Chengjie Wang, Yunsheng Wu, and Dongjin Huang.
\newblock Texdreamer: Towards zero-shot high-fidelity 3d human texture generation.
\newblock \emph{arXiv preprint arXiv:2403.12906}, 2024.

\bibitem[Loper et~al.(2023)Loper, Mahmood, Romero, Pons-Moll, and Black]{loper2023smpl}
Matthew Loper, Naureen Mahmood, Javier Romero, Gerard Pons-Moll, and Michael~J Black.
\newblock Smpl: A skinned multi-person linear model.
\newblock In \emph{Seminal Graphics Papers: Pushing the Boundaries, Volume 2}, pages 851--866. 2023.

\bibitem[Ma et~al.(2024)Ma, He, Cun, Wang, Chen, Li, and Chen]{ma2024follow}
Yue Ma, Yingqing He, Xiaodong Cun, Xintao Wang, Siran Chen, Xiu Li, and Qifeng Chen.
\newblock Follow your pose: Pose-guided text-to-video generation using pose-free videos.
\newblock In \emph{Proceedings of the AAAI Conference on Artificial Intelligence}, pages 4117--4125, 2024.

\bibitem[Moon et~al.(2022)Moon, Choi, and Lee]{moon2022accurate}
Gyeongsik Moon, Hongsuk Choi, and Kyoung~Mu Lee.
\newblock Accurate 3d hand pose estimation for whole-body 3d human mesh estimation.
\newblock In \emph{Proceedings of the IEEE/CVF Conference on Computer Vision and Pattern Recognition}, pages 2308--2317, 2022.

\bibitem[Peng et~al.(2024)Peng, Wang, Zhang, Li, Yang, and Jia]{peng2024controlnext}
Bohao Peng, Jian Wang, Yuechen Zhang, Wenbo Li, Ming-Chang Yang, and Jiaya Jia.
\newblock Controlnext: Powerful and efficient control for image and video generation.
\newblock \emph{arXiv preprint arXiv:2408.06070}, 2024.

\bibitem[Rajasegaran et~al.(2022)Rajasegaran, Pavlakos, Kanazawa, and Malik]{rajasegaran2022tracking}
Jathushan Rajasegaran, Georgios Pavlakos, Angjoo Kanazawa, and Jitendra Malik.
\newblock Tracking people by predicting 3d appearance, location and pose.
\newblock In \emph{Proceedings of the IEEE/CVF Conference on Computer Vision and Pattern Recognition}, pages 2740--2749, 2022.

\bibitem[Rombach et~al.(2021)Rombach, Blattmann, Lorenz, Esser, and Ommer]{rombach2021high}
Robin Rombach, Andreas Blattmann, Dominik Lorenz, Patrick Esser, and Bj{\"o}rn Ommer.
\newblock High-resolution image synthesis with latent diffusion models. 2022 ieee.
\newblock In \emph{CVF Conference on Computer Vision and Pattern Recognition (CVPR)}, pages 10674--10685, 2021.

\bibitem[Rombach et~al.(2022{\natexlab{a}})Rombach, Blattmann, Lorenz, Esser, and Ommer]{Rombach_2022_CVPR}
Robin Rombach, Andreas Blattmann, Dominik Lorenz, Patrick Esser, and Bj\"orn Ommer.
\newblock High-resolution image synthesis with latent diffusion models.
\newblock In \emph{Proceedings of the IEEE/CVF Conference on Computer Vision and Pattern Recognition (CVPR)}, pages 10684--10695, 2022{\natexlab{a}}.

\bibitem[Rombach et~al.(2022{\natexlab{b}})Rombach, Blattmann, Lorenz, Esser, and Ommer]{rombach2022high}
Robin Rombach, Andreas Blattmann, Dominik Lorenz, Patrick Esser, and Bj{\"o}rn Ommer.
\newblock High-resolution image synthesis with latent diffusion models.
\newblock In \emph{Proceedings of the IEEE/CVF Conference on Computer Vision and Pattern Recognition}, pages 10684--10695, 2022{\natexlab{b}}.

\bibitem[Saharia et~al.(2022)Saharia, Chan, Saxena, Li, Whang, Denton, Ghasemipour, Gontijo~Lopes, Karagol~Ayan, Salimans, et~al.]{saharia2022photorealistic}
Chitwan Saharia, William Chan, Saurabh Saxena, Lala Li, Jay Whang, Emily~L Denton, Kamyar Ghasemipour, Raphael Gontijo~Lopes, Burcu Karagol~Ayan, Tim Salimans, et~al.
\newblock Photorealistic text-to-image diffusion models with deep language understanding.
\newblock \emph{Advances in Neural Information Processing Systems}, 35:\penalty0 36479--36494, 2022.

\bibitem[Singer et~al.(2022)Singer, Polyak, Hayes, Yin, An, Zhang, Hu, Yang, Ashual, Gafni, et~al.]{singer2022make}
Uriel Singer, Adam Polyak, Thomas Hayes, Xi Yin, Jie An, Songyang Zhang, Qiyuan Hu, Harry Yang, Oron Ashual, Oran Gafni, et~al.
\newblock Make-a-video: Text-to-video generation without text-video data.
\newblock \emph{arXiv preprint arXiv:2209.14792}, 2022.

\bibitem[Unterthiner et~al.(2018)Unterthiner, Van~Steenkiste, Kurach, Marinier, Michalski, and Gelly]{unterthiner2018towards}
Thomas Unterthiner, Sjoerd Van~Steenkiste, Karol Kurach, Raphael Marinier, Marcin Michalski, and Sylvain Gelly.
\newblock Towards accurate generative models of video: A new metric \& challenges.
\newblock \emph{arXiv preprint arXiv:1812.01717}, 2018.

\bibitem[von Platen et~al.(2022)von Platen, Patil, Lozhkov, Cuenca, Lambert, Rasul, Davaadorj, and Wolf]{diffusers-von-platen-etal-2022}
Patrick von Platen, Suraj Patil, Anton Lozhkov, Pedro Cuenca, Nathan Lambert, Kashif Rasul, Mishig Davaadorj, and Thomas Wolf.
\newblock Diffusers: State-of-the-art diffusion models.
\newblock \url{https://github.com/huggingface/diffusers}, 2022.

\bibitem[Wang et~al.(2023)Wang, Li, Lin, Lin, Yang, Zhang, Liu, and Wang]{wang2023disco}
Tan Wang, Linjie Li, Kevin Lin, Chung-Ching Lin, Zhengyuan Yang, Hanwang Zhang, Zicheng Liu, and Lijuan Wang.
\newblock Disco: Disentangled control for referring human dance generation in real world.
\newblock \emph{arXiv preprint arXiv:2307.00040}, 2023.

\bibitem[Wang et~al.(2018)Wang, Liu, Zhu, Liu, Tao, Kautz, and Catanzaro]{wang2018video}
Ting-Chun Wang, Ming-Yu Liu, Jun-Yan Zhu, Guilin Liu, Andrew Tao, Jan Kautz, and Bryan Catanzaro.
\newblock Video-to-video synthesis.
\newblock \emph{arXiv preprint arXiv:1808.06601}, 2018.

\bibitem[Wang et~al.(2024)Wang, Zhang, Gao, Wang, Zhou, Zhang, Yan, and Sang]{wang2024unianimate}
Xiang Wang, Shiwei Zhang, Changxin Gao, Jiayu Wang, Xiaoqiang Zhou, Yingya Zhang, Luxin Yan, and Nong Sang.
\newblock Unianimate: Taming unified video diffusion models for consistent human image animation.
\newblock \emph{arXiv preprint arXiv:2406.01188}, 2024.

\bibitem[Wang et~al.(2004)Wang, Bovik, Sheikh, and Simoncelli]{wang2004image}
Zhou Wang, Alan~C Bovik, Hamid~R Sheikh, and Eero~P Simoncelli.
\newblock Image quality assessment: from error visibility to structural similarity.
\newblock \emph{IEEE transactions on image processing}, 13\penalty0 (4):\penalty0 600--612, 2004.

\bibitem[Weng et~al.(2023)Weng, Bravo-S{\'a}nchez, and Yeung]{weng2023diffusion}
Zhenzhen Weng, Laura Bravo-S{\'a}nchez, and Serena Yeung.
\newblock Diffusion-hpc: Generating synthetic images with realistic humans.
\newblock \emph{arXiv preprint arXiv:2303.09541}, 2023.

\bibitem[Xu and Loy(2021)]{xu20213d}
Xiangyu Xu and Chen~Change Loy.
\newblock 3d human texture estimation from a single image with transformers.
\newblock In \emph{Proceedings of the IEEE/CVF international conference on computer vision}, pages 13849--13858, 2021.

\bibitem[Xu et~al.(2024)Xu, Zhang, Liew, Yan, Liu, Zhang, Feng, and Shou]{xu2024magicanimate}
Zhongcong Xu, Jianfeng Zhang, Jun~Hao Liew, Hanshu Yan, Jia-Wei Liu, Chenxu Zhang, Jiashi Feng, and Mike~Zheng Shou.
\newblock Magicanimate: Temporally consistent human image animation using diffusion model.
\newblock In \emph{Proceedings of the IEEE/CVF Conference on Computer Vision and Pattern Recognition}, pages 1481--1490, 2024.

\bibitem[Ye et~al.(2023)Ye, Zhang, Liu, Han, and Yang]{ye2023ip}
Hu Ye, Jun Zhang, Sibo Liu, Xiao Han, and Wei Yang.
\newblock Ip-adapter: Text compatible image prompt adapter for text-to-image diffusion models.
\newblock \emph{arXiv preprint arXiv:2308.06721}, 2023.

\bibitem[Yu et~al.(2021)Yu, Zheng, Guo, Liu, Dai, and Liu]{tao2021function4d}
Tao Yu, Zerong Zheng, Kaiwen Guo, Pengpeng Liu, Qionghai Dai, and Yebin Liu.
\newblock Function4d: Real-time human volumetric capture from very sparse consumer rgbd sensors.
\newblock In \emph{IEEE Conference on Computer Vision and Pattern Recognition (CVPR2021)}, 2021.

\bibitem[Zhang and Agrawala(2023)]{zhang2023adding}
Lvmin Zhang and Maneesh Agrawala.
\newblock Adding conditional control to text-to-image diffusion models.
\newblock \emph{arXiv preprint arXiv:2302.05543}, 2023.

\bibitem[Zhang et~al.(2018)Zhang, Isola, Efros, Shechtman, and Wang]{zhang2018unreasonable}
Richard Zhang, Phillip Isola, Alexei~A Efros, Eli Shechtman, and Oliver Wang.
\newblock The unreasonable effectiveness of deep features as a perceptual metric.
\newblock In \emph{Proceedings of the IEEE conference on computer vision and pattern recognition}, pages 586--595, 2018.

\bibitem[Zhu et~al.(2024)Zhu, Chen, Dai, Su, Xu, Cao, Yao, Zhu, and Zhu]{zhu2024champ}
Shenhao Zhu, Junming~Leo Chen, Zuozhuo Dai, Qingkun Su, Yinghui Xu, Xun Cao, Yao Yao, Hao Zhu, and Siyu Zhu.
\newblock Champ: Controllable and consistent human image animation with 3d parametric guidance.
\newblock \emph{arXiv preprint arXiv:2403.14781}, 2024.

\end{thebibliography}
\bibliographystyle{ieeenat_fullname}

\begin{appendices}

\section{Dataset Analysis}

\begin{figure}
\centering
\begin{subfigure}[b]{0.47\textwidth}
   \includegraphics[width=1\linewidth]{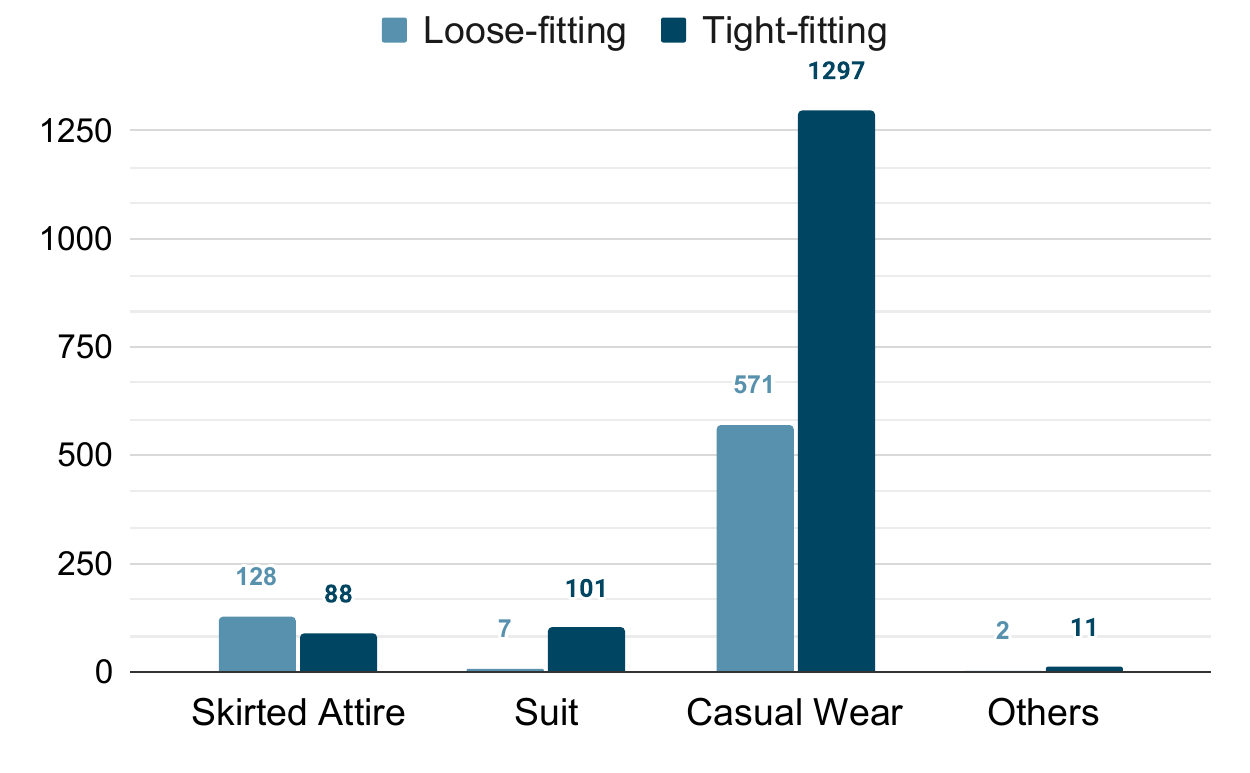}
   \caption{Training data distribution.}
   \label{fig:traindata_dis}
\end{subfigure}
\begin{subfigure}[b]{0.47\textwidth}
   \includegraphics[width=1\linewidth]{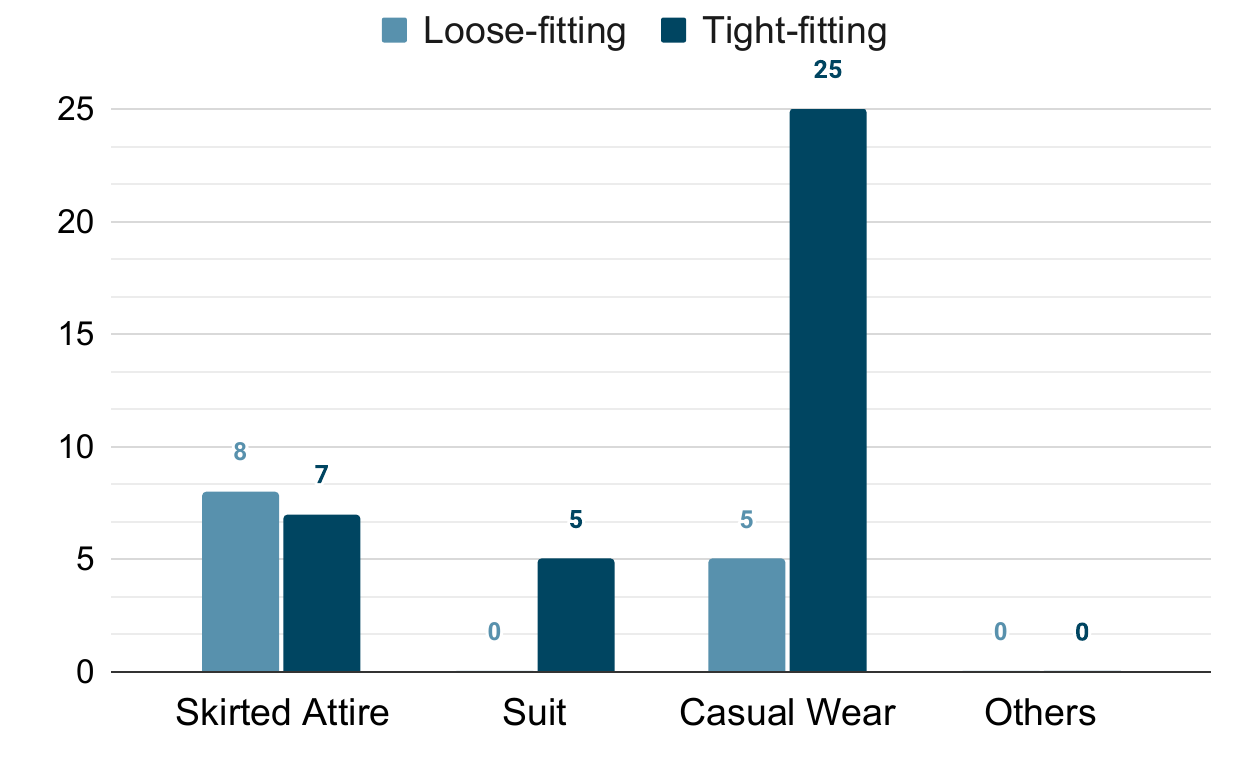}
   \caption{Testing data distribution.}
   \label{fig:testdata_dis}
\end{subfigure}
\caption{Data distribution. We split the clothing type into 4 categories: skirted attire, suit, casual wear, and others. We split each category into two types: loose and tight. We report the number of each category and type and display the overall distribution. We could notice that most clothing is casual wear and a large portion belongs to tight-fitting. }
\vspace{-0.2in}
\end{figure}

Figures~\ref{fig:traindata_dis} and \ref{fig:testdata_dis} present the clothing type statistics of the synthetic training data (2,205 humans) and test data (50 humans). 
 We count people based on four clothing categories: skirted attire, suit, casual wear, and others. In some cases, the clothing belongs to skirted attire and suits or casual wear, we will count this as skirted attire. For each clothing category, we tally two styles: tight-fitting and loose-fitting.

In this paper, we only train on limited human videos, we assume training with more human videos could largely boost the model generalization on the fly. Given that \methodname{} makes use of a cutting-edge 3D pose estimation model and only requires human videos without additional labels for training, it could be trained with numerous and any human videos such as movies, etc.

\section{\methodname{} Training Features}
As has been mentioned in the paper, \methodname{} is in a fully self-supervised fashion. Here we summarize the key training features of our approach:
\begin{itemize}
\item \methodname{} training pipeline (for both stages) is self-supervised.
\item \methodname{} does not use any additional annotations. It is trained with pseudo-ground-truth as we use cutting-edge software which can detect, segment, track and 3Dfy humans (H4D).
\item \methodname{} is scalable and its scaling can be done readily in the future given additional videos of humans in motion and computing resources.
\end{itemize}

\end{appendices}

\end{document}